Goal Conflict in Designing an Autonomous Artificial System

Mark Muraven

University at Albany

# Abstract

Research on human self-regulation has shown that people hold many goals simultaneously and have complex self-regulation mechanisms to deal with this goal conflict. Artificial autonomous systems may also need to find ways to cope with conflicting goals. Indeed, the intricate interplay among different goals may be critical to the design as well as long-term safety and stability of artificial autonomous systems. I discuss some of the critical features of the human self-regulation system and how it might be applied to an artificial system. Furthermore, the implications of goal conflict for the reliability and stability of artificial autonomous systems and ensuring their alignment with human goals and ethics is examined.

Agents can only do one thing at a time. Research suggests, however that people hold multiple conflicting goals (Emmons, King, & Sheldon, 1993; Karoly, 1993). Although this may be problematic at times, there is reason to believe that having multiple conflicting goals may have benefits for autonomous systems as well. In particular, perhaps paradoxically, having conflicting goals may help ensure the stability and reliability of an autonomous system. Having conflicting goals helps to prevents humans from engaging in behaviors that are irrational, perverse, or are self- or other-damaging.

For these reasons, artificial autonomous agents may need to emulate this design. That is, an artificial agent should hold multiple independent goals all competing for completion at the same time. Such a system may be more stable and ultimately may be safer than systems based on other goal structures. With this in mind, this paper has two goals. First, the key features of human goal behavior (self-regulation) will be laid out. This will serve as a brief overview for researchers interested in emulating these features. Second, I will endeavor to explain why such a design is inherently safer, more reliable, and aligned with human interests than other approaches to AI safety.

## Goal Structure

Goals represent desired states that an agent wishes to achieve. A state can be an outcome, event or process (Austin & Vancouver, 1996). People do not need to be aware of their goals to be influenced by them (Custers & Aarts, 2010).  In humans, goals range from biological such as blood oxygenation levels to complex ideas like career success. Similarly, goals can be both short-term (e.g., move a finger) to long-term (e.g., create a new scientific theory).  Goals are **hierarchically arranged** (Cropanzano, James, & Citera, 1993; Powers, 1973). That is, a goal spawns subgoals, which in turn have their own subgoals. Goals at the bottom of the hierarchy are immediate and concrete; as one moves up the hierarchy, goals become more long-term and abstract (Emmons, 1992; Vallacher & Wegner, 1987). People tend to think less about higher level goals but research has shown that people will switch their attention to different parts of their goal hierarchy based on situational feedback (Vallacher, Wegner, & Somoza, 1989).

Given this hierarchical structure, it is clear that a small number of goals must sit atop of hierarchy and must be ultimately responsible for all behavior. The most fundamental of these goals are likely deeply ingrained and unchangeable (Cropanzano et al., 1993; Schwartz, 1992, 1994). People typically may not think about these top-level goals and how they are influencing their behavior (Bargh & Morsella, 2008; Westen, 1998), but they are the ultimate cause of more immediate and limited goals.

There have been numerous theoretical and empirical attempts to catalogue and refine human goals (Austin & Vancouver, 1996; Deci & Ryan, 2000; Murray, 1938; Wicker, Lambert, Richardson, & Kahler, 1984) and a complete list could be quite long. However, when arranged hierarchically, the same general structure seems to emerge from different analytical approaches (Chulef, Read, & Walsh, 2001; Ford & Nichols, 1987; Wicker et al., 1984). This taxonomy may be the most likely candidate for the basic structure of human motivation (see Table 1). The strength of this approach is that the structure fits with theoretical accounts of motivation as well as with ideas of basic human needs (Baumeister & Leary, 1995; Bernard, Mills, Swenson, & Walsh, 2005; Maslow, Frager, Fadiman, McReynolds, & Cox, 1970; Pyszczynski, Greenberg, & Solomon, 1997).

Table 1

Taxonomy of Human Motives

| Broad Cluster | Selected Subclusters |
|---|---|
| Morality and spirituality | Helping others, following values |
| Self-Knowledge/Personal Growth | Gain self-knowledge, pursuit of aesthetics |
| Avoidance | Avoid pain, harm, regret, rejection |
| Relatedness | Making friends, feelings of acceptance, sex, dominance |
| Competence | Gaining control; desire for order and correctness; increasing knowledge and resources |

From Read, Talevich, Walsh, Chopra, and Iyer (2010)

Moreover, similar motives have been identified in infants (Ainsworth, 1969; Brazelton & Greenspan, 2001; Connell, 1990), which further suggests the fundamental nature of these goals. Because they exist in very basic and concrete form in infants, it suggests that these motives may be innate to humans. These initially very primitive and unfocused goals lead to more complex and secondary goals (Deci & Ryan, 1991; Schaffer & Emerson, 1964; Thelen & Smith, 1996). Identification of the processes underlying this development is beyond the scope of this manuscript, but there is likely some level of **habituation**, where if the same goal-directed behavior is repeatedly encountered, it loses impact (Gerson & Woodward, 2014). Thus, over time, the system looks for new means to fulfill goals. Humans also learn goals from direct reinforcement, classical conditioning (Berridge, 2001) and perhaps most importantly, from watching others (Hamlin, Hallinan, & Woodward, 2008; Luo & Johnson, 2009). Indeed, even at an early age, babies copy what they see others do and by doing so may learn goal structure (Woodward, Sommerville, Gerson, Henderson, & Buresh, 2009).

Critically, because these goals are independent and unrelated, they may often be in conflict (Emmons et al., 1993; Karoly, 1993). One could rescue a drowning person by either swimming out to them or throwing a rope but both are not possible. Even more fundamental goals may conflict as well. The desire to explore may directly contradict the desire for safety. One may search for compromise solutions that minimize the degree of conflict, but ultimately these two fundamental desires cannot be reconciled. One must choose one path or another. As described below, although conflict among goals may seem like an undesirable feature in an autonomous system, such conflict has distinct advantages for an autonomous system and may be critical to the stability and reliability of the system in complex and changing environments.

## Goal Selection

Obviously, this conflict between goals can be debilitating to a system. For example, progress toward one goal may lead to a greater discrepancy for the other. The system can then respond to putting resources into the other goal, thereby undoing the progress made toward the first goal. Hence, the system believes that it is maintaining balance when, in fact, it is simply wasting resources. Thus, a system needs extensive control processes to ensure that goal conflicts are resolved in a resource positive way (Powers, 1973). Even basic systems can get paralyzed by conflicting choices and hence there needs to be a way to resolve conflicts (for an example of how an autonomous sytem might reject immoral commands, see Briggs & Scheutz, 2015).

First, there must be a system in place to select one goal and inhibit all others. Cognitive scientists have highlighted the role of executive functions in this process (Hofmann, Schmeichel, & Baddeley, 2012; Shallice & Burgess, 1993). That is, people have means to suppress competing desires and goal (Fishbach & Shah, 2006; Shah, Friedman, & Kruglanski, 2002). Consistent with that argument, to the extent that people have problems with their executive functioning, they have problems maintaining goal directed behavior (Mischel, Shoda, & Rodriguez, 1989; Stroebe, Mensink, Aarts, Schut, & Kruglanski, 2008).

A deeper question that is more relevant to the design of an artificial autonomous agent is why a particular goal gets selected and why the other goals are suppressed. A good number of critical variables have been identified in this process. One of the most prominent is **expectancy-value theory** (Ajzen, 1991; Atkinson, 1964; Wigfield & Eccles, 2000). This theory follows from the economic utility idea that people choose to work on tasks based on the value they assign to the outcome and their ability to reach that goal. Value may arise from prior rewards or punishments but may also be affected by intrinsic interest, cost of action, and utility of the behavior. Outcomes that are more valued are more likely to be pursued, although expectation of success moderates that process.

Despite the overall usefulness and ubiquity of expectancy-value theory, other factors may also influence goal selection. For instance, **habit**, has been found to be a very strong predictor of goal directed behavior (Aarts & Dijksterhuis, 2000; Wood & Neal, 2007). If a goal has been selected in the past, it is more likely to be selected in the future. This may be automatic (Aarts & Dijksterhuis, 2003), to save cognitive effort  but may also represent the fact that prior goals are likely to be the most important, as well (Ronis, Yates, & Kirscht, 1989).

Situational **priming** also has been found to play a role in goal selection. Goal selection is likely affected by environmental and social cues (Bargh & Ferguson, 2000; Gollwitzer & Brandstätter, 1997). Cues in the environment may activate goal constructs. The goal itself may also activate related goals (Anderson, 1996; Collins & Loftus, 1975; Posner & Snyder, 1975) as well as inhibit competing goals (Fishbach, Friedman, & Kruglanski, 2003). This helps to ensure that the system is responsive to changing circumstances and provides flexibility to goal selection.

Outcomes that are perceived as closer in space, time, or psychological distance are also valued more than more distal outcomes, based on the **construal** process (Fujita, Trope, Liberman, & Levin Sagi, 2006; Trope & Liberman, 2010). Likewise, from the **framing** research, losses weigh heavier than gains (Förster, Higgins, & Idson, 1998; Roney, Higgins, & Shah, 1995). Thus, goals that are perceived to result in a loss are avoided as opposed to goals that result in a gain. These theories suggest that features of the goals play a role in their selection (Yang, Stamatogiannakis, & Chattopadhyay, 2015).

Finally, outcomes that fulfill multiple goals may be valued more than outcomes that only fulfill one goal (**multifinal**). Although people look for ways to minimize goal conflict through the pursuit of several goals at once, such compromise may be rare and may be less efficient than pursuit of a singular goal (Köpetz, Faber, Fishbach, & Kruglanski, 2011; Kruglanski et al., 2002). For instance, studying in a group may fulfill both achievement and social goals, but may be less effective than studying alone. In general, it seems that people will choose multifinal goals when possible, but when one goal is highlighted, multifinal goals are less likely to be selected (Fishbach & Zhang, 2008; Zhang, Fishbach, & Kruglanski, 2007).

There have been some attempts at unifying these different approaches into a coherent model. Most notably, Ballard et al.'s (2016) extended Multiple-Goal Pursuit Model (MGPM*) was able to successfully predict decision making under varying circumstances. Further refinements may lead to even better models of goal selection in multiple-goal situations that integrates more of these approaches

## Goal Switching

The selection of a goal does not mean that goal will be pursued single-mindedly to its conclusion. Indeed, there is considerable evidence that people will revisit the goal selection process and change course (Baumgartner, Pieters, Haugtvedt, Herr, & Kardes, 2008; Wang & Mukhopadhyay, 2012). Although goal switching may seem like a negative feature, it helps ensure that the system remains responsive to changing situations, pays attention to its own impact on the world, and avoids irrational persistence. Thus, even pleasurable and goal directed activities, like socializing with friends, should become a lower priority and will be interrupted by other desires after a while.

Clearly, goal switching will occur when a goal is **satisfied**. Once a goal is met, the system may begin the search for a new goal. Because, as noted above, higher order goals are long-term and abstract (and therefore complete satiation is not feasible), the system likely focuses on intermediate level goals (e.g., pass a test) and their completion rather than either high-level (e.g., career success) or low-level (e.g., turn the page in a book). In this way, goal progress may be more closely monitored than actual goal completion (Carver & Scheier, 2012; Simon, 1967) with people flexibly adjusting their effort exerted toward a given goal without necessarily abandoning the goal altogether.

The search for a new goal may also happen when a goal is **blocked**. If insufficient or too slow progress is being made (which implies the system has some sense of the desired rate of progress), the system should try to find a new goal (Dreisbach & Goschke, 2004; Fung & Carstensen, 2004; Simon, 1967). For some goals, in particular lower level goals in service to higher priority goals, the search might focus on ways around that obstacle, rather than dropping the higher level goal altogether (Geers, Wellman, & Lassiter, 2009; Markus & Wurf, 1987; Wrosch, Scheier, Miller, Schulz, & Carver, 2003). For example, if the higher-level goal is to help an injured person, but the door is locked, the system should look for alternative ways to enter the room before dropping the goal of helping the person altogether. Research has found that people rarely drop high level, person defining goals, but there may be times when it is clear that a goal is not feasible (e.g., becoming a doctor after a low score on the MCAT) that the person has to revisit them (Cropanzano, Citera, & Howes, 1995; Donovan & Williams, 2003).

Less dramatically, but perhaps more importantly, goal switching may happen in response to **situational cues**. There is reason to believe that people continue to monitor their environment, even while working on a goal and if a cue signals the availability of a high priority goal, switching may occur (Arrington, Weaver, & Pauker, 2010; Custers & Aarts, 2010). For example, the chime of a new message on a phone might lead an author to stop writing, even if writing is very important to them. Such task switching may be viewed as a liability, but may serve a valued function to ensure that the system is responsive to changing situations.

Finally, even when adequate goal progress is being made and there is no strong signal to change goals, the system seems to have a built-in **fatigue** process. The continued suppression of competing goals should weaken over time, so that there is an increasing likelihood of switching goals the longer and more energetically they are suppressed (Hagger, Wood, Stiff, & Chatzisarantis, 2010; Muraven, 2012). This should be true for any task, not just tasks that are unpleasant or undesirable. Again, such failures of self-control may be seen as a problem in humans, but it has been argued that such a fatigue system is necessary to ensure proper goal functioning (Inzlicht, Schmeichel, & Macrae, 2014). Without a fatigue process, a system may pursue goals to excess; the fatigue process helps to ensure a better balance in goal pursuit. Obviously, there may be a complex feedback system, so the fatigue process itself (like the response to situational cues) can be modified by internal and external signals so that the system may be able to ignore fatigue feedback temporarily, perhaps at the cost of greater eventual fatigue (Muraven, Shmueli, & Burkley, 2006; Muraven & Slessareva, 2003).

## Strengths of Approach

As described, an artificial autonomous system with multiple conflicting independent goals clearly has some significant limitations in its ability to successfully pursue goals. It is prone to distraction, giving up, and may be less efficient. However, these weaknesses need to be weighed against several significant strengths. First, because it is modelled on an autonomous system that is well-established and proven to work (i.e., humans), we gain some confidence that it may work for artificial systems as well.

Beyond that, the checks and balances built into the system help to ensure its long-term stability and reliability. For example, the goal switching process ensures that an artificial autonomous agent will eventually escape local minimums or obsessive type behavior. Indeed, the entire conflict resolution system avoids many of the problems associated with misspecified goals. There simply is not one goal to misspecify and if one of the multitude of goals is poorly described, the eventual impact on the overall system will be reduced.

Because the starting point is goals that humans hold, the system is innately aligned with human goals. Thus, there should be less concern that the autonomous agent will pursue goals that go against what humans' value. Due to the embodied nature of the goal training, it is difficult to ensure the complete reliability of the system, however. Although it may be possible to overtrain certain motives (e.g., never hurt a human), this may have negative repercussions for the overall system (e.g., it may find it impossible to give an injection).

Because the goals are embodied and the complex interplay among the goals is what drives autonomy, any effort to change or remove a goal after training is likely to result in catastrophic failure of the system. Complexity arises from development, but the underlying goal still remains—the desire for feelings of competence that drives a baby's exploration is what makes an adult start a hobby, even if the behavior is more complex and links to the deeper motive are not readily apparent. Thus, the system is highly resistant to change by either external or internal forces.

Although this bottom up approach avoids the need to specify ethical rules, such rules may be permitted in the conflict resolution module. That is, some motivational goals may be sacrosanct and override all other rules. Such ethical standard can be programmed in to be innate, or, more likely, be taught as part of the training process and deeply ingrained in the goal structure. Infants do not innately know not to eat feces, but to most adults, the thought never even enters the mind and rejected out of hand if suggested. Thus, the learning of ethical rules is part of the training process.

# Recommendations and Future Directions

Given the strength of this approach to designing autonomous artificial agents, there are several lines of research that I recommend pursuing. As described above, people hold multiple goals that independently influence behavior and the conflict between these goals needs to be resolved using a complex algorithm. This suggests that goals should be modeled as independent agents co-existing in one decision-making body. Such multi-agent systems have garnered increasing interest in recent years. Further research into the design of such systems may be warranted. In particular, understanding how such systems resolve conflict and work together may help in the design of autonomous artificial agents (see, e.g., Leibo, Zambaldi, Lanctot, Marecki, & Graepel, 2017).

Furthermore, the training of multi-agent systems is complex but a necessary step to create the independent goal model outlined here. Humans likely have certain innate goals that get reinforced through development. This suggests that goals themselves are embodied. Understanding how to train such independent goals in a multi-agent system and increase the complexity of the learned goals is an area that requires more research (Mnih et al., 2015; Zawadzki, Lipson, & Leyton-Brown, 2014). It is likely that a combination of classical and operant conditioning, as well as social learning will be necessary.

From the cognitive science side, more complex and complete models of goal conflict are necessary. There are models that describe goal selection based on a number of internal and external factors (e.g., Ballard, Yeo, Loft, Vancouver, & Neal, 2016) but there are still more factors to include. Moreover, the conflict resolution system likely contains complex feedback systems, so that it may modify itself. Models of goal switching are less developed and thus also need more formal description.

Finally, the verification and validation of this independent goal model presents significant challenges. As a start, it might be useful to determine how such a system may resolve well known goal conflicts (such as the trolley problem). Such tests can be conducted with simplified models now (Arnold, Kasenberg, & Scheutz, 2017) . However, as outlined above, the goal structure is embodied and thus every system may need its own verification process to ensure the reliability of the training. Creating tests to verify and validate multi-agent models may be a necessary step to ensuring the reliability of any goal training.

In conclusion, an artificial autonomous agent may be best modeled by a self-regulating system that consists of multiple independent and conflicting hierarchical goal structures. Although complicated and difficult to design and verify, has several noteworthy strengths in ensuring long-term stability and human alignment. The system can also grow in complexity over time, so that it can be scaled up or down, depending on the situational needs. It can also be individualized, based on both goals activated and trained, to fit varying circumstances. In short, this may be one way to move forward toward an artificial autonomous system.